\documentclass[conference]{IEEEtran}
\IEEEoverridecommandlockouts
\usepackage{cite}
\usepackage{amsmath,amssymb,amsfonts}
\usepackage{algorithmic}
\usepackage{graphicx}
\usepackage[section]{placeins}
\usepackage{textcomp}
\usepackage{xcolor}
\usepackage[ruled,vlined]{algorithm2e}
\usepackage{multirow}
\usepackage{array}
\usepackage{caption}
\usepackage{subcaption}
\usepackage{tikz}
\usetikzlibrary{spy, calc}
\usepackage[acronym]{glossaries}

\newacronym{som}{SOM}{Self-Organizing Map}
\newacronym{cnn}{CNN}{Convolutional Neural Network}
\newacronym{lgnn}{LGNN}{Locality Guided Neural Network}
\newacronym{lime}{LIME}{Local Interpretable Model-agnostic Explanations}
\newacronym{xai}{XAI}{Explainable Artificial Intelligence}

\def\BibTeX{{\rm B\kern-.05em{\sc i\kern-.025em b}\kern-.08em
    T\kern-.1667em\lower.7ex\hbox{E}\kern-.125emX}}
\begin{document}

\title{Locality Guided Neural Networks for Explainable Artificial Intelligence\\}

\author{\IEEEauthorblockN{Randy Tan, Naimul Khan, and Ling Guan}
\IEEEauthorblockA{\textit{Department of Electrical and Computer Engineering} \\
\textit{Ryerson University}\\
Toronto, Canada\\
randy.tan@ryerson.ca, n77khan@ryerson.ca, lguan@ryerson.ca}
}

\maketitle

\begin{abstract}

In current deep network architectures, deeper layers in networks tend to contain hundreds of independent neurons which makes it hard for humans to understand how they interact with each other. By organizing the neurons by correlation, humans can observe how clusters of neighbouring neurons interact with each other. In this paper, we propose a novel algorithm for back propagation, called \gls{lgnn} for training networks that preserves locality between neighbouring neurons within each layer of a deep network. Heavily motivated by \gls{som}, the goal is to enforce a local topology on each layer of a deep network such that neighbouring neurons are highly correlated with each other. This method contributes to the domain of \gls{xai}, which aims to alleviate the black-box nature of current AI methods and make them understandable by humans. Our method aims to achieve \gls{xai} in deep learning without changing the structure of current models nor requiring any post processing. This paper focuses on \glspl{cnn}, but can theoretically be applied to any type of deep learning architecture. In our experiments, we train various VGG and Wide ResNet (WRN) networks for image classification on CIFAR100. In depth analyses presenting both qualitative and quantitative results demonstrate that our method is capable of enforcing a topology on each layer while achieving a small increase in classification accuracy.

\end{abstract}

\glsresetall

\section{Introduction}
There has been significant progress in deep learning in the past few years, especially with recent advances in computational power and the emergence of large datasets to train models. Deep learning techniques have taken the top position as state of the art in many domains such as image processing \cite{rawat_deep_2017} and natural language processing \cite{young_recent_2018}. The development of software libraries have simplified the process of training networks to the point where domain experts that only have a small amount of deep learning knowledge can build models. However, one major challenge for deep learning is its inherent black box nature, in which users have little control or knowledge on what type of features are learned in the hidden layers. This lack of knowledge can prohibit the application of deep learning in critical domains; where misclassifications have a high cost. Until we can explain how deep learning models have come up with a decision, it will be difficult to apply them in areas such as self-driving cars, medical diagnoses, and other critical domains \cite{noauthor_general_nodate}. Another application where it is important to understand what type of semantic information a model learns is in transfer learning \cite{wang_videos_2018}. Many of these methods take pretrained networks that are trained on large datasets like ImageNet \cite{deng_imagenet_2009} or Kinetics \cite{carreira_quo_2017} and then utilize only the feature extraction portion of the model and feed it through a different classifier. In these types of methods, the classifier is forced to treat all of its input features as independent and unknown features due to the blackbox nature of deep learning. If domain experts could understand the features from deep learning methods, they may be able to apply their own knowledge and form handcrafted features that could improve the classifier.        

One challenge in trying to understand neural networks is that the neurons within a given layer are independent of each other. In current deep learning structures, neurons are only connected to other neurons in the previous and next layers; none of the neurons within the same layer are connected to each other. This makes it difficult to find any inherent relationship between features without comparing all of the neurons in the same layer together. We propose a new algorithm for back propagation inspired by \gls{som}, that can enforce a topology on each layer where neighbouring neurons learn similar concepts. While in this work we focus on convolutional neural networks with images, this method is theoretically applicable to other networks types. Enforcing a topological structure that gathers similar filters together will make it easier for humans to visually understand the deeper layers of a network. The work by Google \cite{olah_building_2018} identified the concept of \textit{Neuron Groups} in which certain visual concepts activate specific groups of neurons. Our method aims to group these neurons together in the topology. The primary contributions of our \gls{lgnn} method include:
\begin{itemize}
    \item Clustering neurons in each layer of a network using a similar neighbourhood function to \gls{som} such that neighbouring filters share semantic concepts.
    \item Only modifying the gradient update step for convolutional layers such that it can be integrated into any state of the art \gls{cnn} model with negligible computational overhead. 
    \item Enforcing the topology during training without any post-processing, unlike methods like \cite{kim_interpretability_2018, bau_network_2017}
    \item Slightly increasing accuracy for image classification due to the regularizing effect of neighbourhood functions
\end{itemize}

\section{Background}\label{background}
\subsection{Explainable Artificial Intelligence} \label{back_understandding}

The \gls{xai} program was started by DARPA in their interest for autonomous systems that can make critical decisions. While they credit machine learning methods for their high success, they state their main limitation is their inability to explain their decisions to human users. The main questions they want answered by machine learning systems include: "Why did you make your decision?", "When do you succeed or fail?", and "How can I correct your errors?". The target for the \gls{xai} program is for a new generation of machine learning methods that can have their thought process explained to human users. Researchers have mainly targeted the problem of \gls{xai} in two ways: 
\begin{itemize}
    \item Post hoc methods that can attach explanations to already trained models
    \item Modifying or creating new machine learning models that are more inherently understandable by humans
\end{itemize}

Research in post hoc interpretation of machine learning have tried to tackle the problem from several angles. One example is \gls{lime} which ignores the AI model and explains the relationship between a single input and its output by approximating a small local area around the input as a linear classifier \cite{ribeiro_why_2016}. For \gls{xai} in \glspl{cnn}, some recent works try to tie neuron activations to semantic concepts humans can understand. One method is Network Dissection \cite{bau_network_2017}, which is a framework that assigns quantitative values to how interpretable an individual neuron is to a semantic concepts such as scenes, objects, textures, and colors. Another method is Testing Concept Activation Vectors (TCAV) \cite{kim_interpretability_2018}, which translates semantic concepts to a vector within a layer's output space rather than an individual neuron. TCAV finds the vector by first creating a training set consisting of images that contain a concept and images that do not. The images are fed through the network and the activations at the desired layer are used as an input to a linear classifier. The vector orthogonal to the linear boundary is used to represent the concept.

Several \gls{xai} methods, including our method, propose ways to modify models to be inherently more understandable. Variational Auto Encoders (VAE) and their variants are a good example of a specific network model that is explainable \cite{kingma_stochastic_2014, fortuin_som-vae_2019, esser_variational_2018}. Autoencoders consist of an encoder that compresses an input to a latent space, and a decoder that restores the original input vector. VAEs modify the encoder such that it returns a probability distribution instead of a single latent variable and also adds a regularization term to the loss such that the latent space becomes continuous, where vectors close to each other in the latent space have similar appearances in the original input space. This allows users to sample the latent space and still get coherent decoded outputs. In one \gls{xai} application, \cite{wang_videos_2018} proposed a network architecture that inherently learns action recognition in an explainable way. Rather than classifying the action directly, their method detects objects first. The objects are then fed through a Graph Connected Network (GCNs) \cite{kipf_semi-supervised_2017} to determine how the objects interact to classify actions. Since the network has been broken up into steps, a human can see what types of objects were detected by the first step, and also see what type of connections the second step deemed as important for classification. 

Our method explores \gls{xai} by simply re-organizing the order of filters such that users can have a more global understanding of each layer compared to looking at the unordered filters individually. In the ideal scenario, the filters trained by \gls{lgnn} would be identical to the baseline, but organized such that the filters that all activate on similar semantic concepts are gathered together. In \cite{olah_building_2018}, the authors identified the groups of filters that activated on similar concepts as \textit{Neuron Groups}, but their work did not use these groups to reorganize the filters. 

\subsection{Network Visualization}\label{back_visualization}

Since \glspl{cnn} are applied to the domain of image processing, it is natural that many methods try to explain \glspl{cnn} through visualization. Two categories of visual \gls{cnn} interpretations include attribution/saliency and feature visualization \cite{simonyan_deep_2014, olah_feature_2017}. Attribution methods show what parts of an example image triggered a specific neuron activation \cite{selvaraju_grad-cam:_2019, zeiler_visualizing_2013, springenberg_striving_2015, olah_building_2018}. They first take an input image and feed it through a trained \gls{cnn}. They then try to map the neuron response back to the pixel regions in the original image that activated it. The issue with attribution methods is that they only show a correlation between an input image and neuron activation. In traditional \glspl{cnn}, individual neurons at the higher levels can represent mixtures of patterns or concepts \cite{zhang_interpretable_2018}. Since attributions are tied to dataset examples, they could possibly mislead users by only showing a portion of what a neuron is looking for. 

Feature visualization methods try to numerically generate a new image that maximizes a specific neuron's activation. They start with a noise image and trains the image to maximally activates a neuron. It accomplishes this by setting the training loss to the negative activation of the desired neuron, freezing the network weights, and passing the gradients all the way to the noise image. The resulting image should be a pattern that represents what the desired neuron is looking for. There are several regularization techniques listed by \cite{olah_feature_2017} that can help make clearer visualizations including; high frequency penalization, transformation robustness, and learned priors.

\subsection{Self Organizing Maps} \label{back_som}

This work uses a locality enforcing algorithm that is motivated by \gls{som} \cite{kohonen_essentials_2013}. \gls{som} is an unsupervised clustering algorithm that is often compared to Vector Quantization (VQ) or K-Nearest Neighbours (KNN). The unique feature of \gls{som} compared to other clustering methods is that \gls{som} sorts its clusters by similarity. Section \ref{prop_lgnn} explains in fuller detail, but essentially \gls{som} achieves the locality based ordering by defining a structure (typically a 2-D grid) for the clusters during initialization as a set of connected neighbours for each node and then propagating any gradients that each node receives to its neighbours.

Several works including \cite{kyan_approach_2015}, \cite{fortuin_som-vae_2019} and  \cite{wang_deep_2017} are examples of methods that have employed \gls{som} in AI. In \cite{kyan_approach_2015}, Spherical Self Organizing Maps (SSOM) is used as an extension of \gls{som}, to evaluate ballet dance performance. The only difference between \gls{som} and SSOM is that SSOM is structured as a tessellated enclosed sphere as opposed to a 2D grid. This work treated individual neurons within the SSOM as a posture and classified full ballet performances as a trajectory through the SSOM. In \cite{fortuin_som-vae_2019}, the main goal of the work was to take the high dimensional features from deep networks and to make them more visually interpretable for humans; specifically for time series. The work used a VAE as the baseline model and further compresses the encoding by representing it as a neuron within a trained \gls{som}. The main contribution in the paper was the introduction of several loss functions, which allow the backpropagation to overcome the non-differentiability of discretization and for the \gls{som} to be traversed smoothly in time. Similar to \cite{kyan_approach_2015}, this method can encode a time series as a trajectory through a \gls{som}. In \cite{wang_deep_2017}, a \gls{som} was trained on the FC layer at the end of a deep network as a means of quantization for Aproximate Nearest Neighbour (ANN) search. Their method leveraged the topology preserving capabilities of \gls{som} to train a quantization loss along side the classification loss where similar image pairs minimized their distance on the \gls{som} map while dissimilar images maximized their distance. By attaching \gls{som} to the end of a deep network, they were able to acquire input features from the FC layers that contain information about the image. Overall, their method is capable of both ANN search and classification. 

\section{Proposed Method}\label{prop}

\subsection{Locality Guided Neural Networks} \label{prop_lgnn}

The main goal of this work is to interfere as little as possible with current \gls{cnn} structures while allowing correlation along the channel dimension to be inherent. Our method will attempt to enforce a topology on each layer such that neurons that search for similar concepts are clustered together. This will make it easier for users to have a more global understanding of each layer rather than looking at individual neurons. Additionally, having locality within the learning will allow information to be propagated within each layer instead of only propagating between layers. 

Our algorithm borrows its topology preserving properties from \gls{som}. In the original \gls{som} algorithm, neurons are arranged into a 2D grid and contain a location in the grid $l_j$ and a weight $w_j$ for each neuron $j$. The weight corresponds to the position of the node in the high dimensional input space. The location of the neuron would be its position in the 2D latent space. \gls{som} employs competitive learning such that during each iteration, only a single neuron wins a gradient from the input. \gls{som} also employs a neighbourhood function such that after a neuron wins, a portion of the gradient is passed to its neighbouring neurons in the latent space. When fully trained, the competitive learning and the neighbourhood functions cause the \gls{som} map to behave like a blanket being spread on a higher dimensional surface. 

At the beginning of the \gls{som} algorithm, the grid is initialized with random weights. Each iteration, a single sample is selected from the input and the node with the weight that has the smallest Euclidean distance is selected as the winning node:
\begin{equation}
    c = argmin_j (||x[t]-w_j[t]||^2)
    \label{eq:somerror}
\end{equation}
where $c$ is the index of the winning node, $x[t]$ is the input at iteration $t$, and $w_j[t]$ is the weight of neuron $j$. The winning node and its neighbours receive a gradient that is inversely proportional to its distance in the latent space to the winner:

\begin{equation}
    w_j[t+1] = w_j[t] + \alpha [t] (\eta_{cj}[t] (x[t]-w_j[t]))
    \label{eq:somupdate}
\end{equation}

where $\eta_{cj}[t]$ is the neighborhood function and $\alpha [t]$ is the learning rate. The neighborhood function is a monotonic decreasing function centered on the winning node. In this work we use a Gaussian window of a fixed size and shrinking $\sigma$ as our neighbourhood function. Overall, this means that the winning node receives a large pull towards the input, while the neighbouring nodes receive a pull of diminishing strength as the nodes get further from the winner. By having the winner pass a portion of the gradient to its neighbours, the neighbours all start to share similar semantics after a few iterations.  

While \gls{som} is a type of neural network, its structure is not meant for deep learning. Previous works such as \cite{fortuin_som-vae_2019} and \cite{wang_deep_2017} have already combined \gls{som} with deep learning models as explored in Section \ref{back_som}, but in both of the works \gls{som} was trained on the output of a single layer. In this work however, we wish to enforce a SOM-like locality on each layer of a \gls{cnn}. That way, the filters of each layer can be arranged along the channel dimension and have its topology preserved. Since the filters of a \gls{cnn} already receive gradients from back propagation, we decided to use those gradients rather than acquiring them from the input like in traditional \gls{som}. Additionally, while we initially tried to incorporate competitive learning to our method, we found during our initial test that it significantly reduced the rate at which the network was learning and so it was taken out. In a future work, competitive learning could be added back in. Since the only part of \gls{som} we are keeping is the neighbourhood function, we call our method \gls{lgnn} to avoid confusion.

Before the weight update function from \gls{som} can be used, it needs to be modified to fit the gradients from deep learning. In \glspl{cnn}, the gradients now come from back propagation and are accumulated over batches and so we denote $g[k]$ as the gradient of the $k^{th}$ filter which has been accumulated over a batch of images. We also note that since we have a batch of accumulated gradients instead of the gradient of a singular winner of an iteration, the neighbourhood function needs to shift and be applied to each of the gradients. From equation \ref{eq:somupdate}, we remove the time arguments and add a summation over $k$ accumulated gradients in a batch. The new weight update function for each batch is:

\begin{equation}
    \begin{split}
        w_{new}[j] &= w_{old}[j] + \alpha \sum_k (\eta [j-k] g[k]) \\
        &= w_{old}[j] + \alpha (\eta \ast g) [j]
    \end{split}
    \label{eq:lgnnupdate}  
\end{equation}

As shown in equation \ref{eq:lgnnupdate}, the term on the right is equivalent to performing a convolution between the accumulated gradients and the neighbourhood function. In other words, to achieve locality between filters, we simply reshape the channel dimension of the gradient matrix and then low pass filter it. The pseudocode for \gls{lgnn} is as shown in Algorithm \ref{alg:lgnn}. The product of the \gls{som} dimensions $[m,n]$ need to be equal the number of filters in the layer. Since each individual layer can have different number of output channels, we store the \gls{som} dimensions as a lookup table. 

\begin{algorithm}
\DontPrintSemicolon
\emph{$c_{out} =$ output channels, $c_{in} = $ input channels, $s =$ filter size, $[m,n] = $ \gls{som} dimensions}\;
\Begin{
    Initialize network\;
    Initialize LPF with fixed weights\;
    \For{each iteration}{
        Zero gradients\;
        Input batch into network and calculate gradients from back propagation\;
        \For{each layer}{
            Reshape gradient tensor: $[c_{out}, c_{in}, s, s] \xrightarrow{} [(c_{in} \times s \times s), 1 , m, n]$\;
            Apply LPF\;
            Reshape tensor back: $ [(c_{in} \times s \times s), 1 , m, n] \xrightarrow{} [c_{out}, c_{in}, s, s]$\;
        }
    Apply gradients
    }
}
\caption{Locality Guided Neural Networks\label{alg:lgnn}}
\end{algorithm}

For our neighbourhood function, different LPF sizes were tested. It was found that when the \gls{lgnn} switched from a $3 \times 3$ LPF to a $5 \times 5$ LPF, the validation accuracy dropped by more than $2\%$. We suspect that this is due to the fact that \gls{lgnn} gets its gradients from back propagation rather than having the neurons matching the input distribution. While for \gls{som} there is a singular input distribution, there are several local minimums that a network can settle on. If the pulling effect from the neighbourhood function is too strong, it may force the network to overfit to the training set and not generalize as well to new data.

There are several benefits to \gls{lgnn}. Firstly, this method allows information to propagate locally within a layer. Normally, information in neural networks only propagates between layers, which can allow neurons within the same layer to carry redundant or conflicting information. The second benefit to \gls{lgnn} is that applying it to a network only modifies the accumulated gradients during back propagation. No modifications are done to the network configuration nor the forward pass, and there is no required post-processing. Since back propagation is the only thing changed, an \gls{lgnn} version of a network will have identical inference time to the baseline model. Also, since the accumulated gradients are the only tensors affected by \gls{lgnn}, the added computational time of \gls{lgnn} does not scale with image or batch size since the gradients are already accumulated for an entire batch. 

\section{Experiments and Results} \label{exp}

\subsection{\gls{lgnn} Applied to VGG and WRN} \label{exp_app}

In our first experiment, we compare the performance of \gls{lgnn} against their baseline models. For these experiments, CIFAR-100 is used as the dataset. We use two different baseline models: VGG \cite{simonyan_very_2014} and Wide ResNet (WRN) \cite{zagoruyko_wide_2016}. For VGG, we use VGG-11 and VGG-19. For these models, we replace the last 3 FC layers with a single FC layer as \cite{springenberg_striving_2015} demonstrated that multiple FC layers adds too many network parameters without much benefit. For WRN, we use WRN-16-8 and WRN-28-10. All models use batch normalization \cite{ioffe_batch_2015} and the WRN models use the dropout version with 0.3 dropout. The optimizer is SGD with a momentum of 0.9 and a weight decay of 0.0005. VGG-11 ran for 100 epochs with an initial LR of 0.1 and decreased by a factor of 0.2 at epochs (40, 70, 90). The other 3 networks ran for 200 epochs with initial learning rate of 0.1 and decreased by a factor of 0.2 at epochs (60, 120, 160) just like in \cite{zagoruyko_wide_2016}. The reason why VGG-11 has different hyper parameters is because the model was small enough to finish training at 100 epochs and we wished to keep training time short for this model in order to test the different versions. For the neighbourhood function, a $3 \times 3$ Gaussian LPF was used with a $\sigma$ of $0.5$. Replication padding was used whenever the LPF was applied. In most of our experiments, two different version of the neighbourhood function were tested; in the first version the LPF stayed the same the whole time and in the second version, $\sigma$ was changed at the rate of:

\begin{equation}
   \sigma =  0.5 \times (1- \frac{current \;\; epoch}{total \;\; epochs})
    \label{eq:sigmadecay}
\end{equation}

In the results, we call the trials either constant or decreasing to refer to $\sigma$. Since \gls{lgnn} does not change the network configuration or number of parameters, we were able to save 5 sets of random initializations and then ran each network configuration on each of those saves. When looking at the final results, we use the median instead of the mean as it is more robust against outliers when the number of trials is low. 

\begin{table}[htbp]
    \centering
    \begin{tabular}{|c|c|c|c|}
          \hline
           VGG size & Regular & LGNN-Constant & LGNN-Decreasing \\
          \hline
          \hline
          VGG-11& 69.96\ & 70.16 & 70.17 \\
          \hline
          VGG-19& 72.01 & 72.39 & 72.46 \\
          \hline
    \end{tabular}
    \caption{Accuracies (\%) for VGG}
    \label{tab:vgg}
\end{table}

The results for VGG-11 and VGG-19 are shown in table \ref{tab:vgg}. For the VGG networks, the median performances for all 3 networks are not dramatically different, with both versions of \gls{lgnn} pulling slightly ahead of the baseline version. Also, having a decreasing or constant sigma did not significantly affect the median performance. Between the two sizes, VGG-19 had a larger performance increase. While the performance difference is minimal, we have demonstrated that training a network to have a local topology did not adversely affect the performance of the network.   

Wide ResNets have a more complex structure than VGG with various convolution filter sizes and residual branches. Therefore, we test \gls{lgnn} on various combinations of layers. In ResNets, the shortcut path is supposed to grant an easier path for gradients to flow to lower layers and so there is a possibility that applying \gls{lgnn} to the $1 \times 1$ convolutions in them may decrease performance. Also, the very first convolution in WRN contains far fewer filters than the other layers and applying locality to such a small number of filters may harm the network's ability to learn distinct features. Therefore, the first configuration we use only applied \gls{lgnn} to the main branch of the resblocks without changing the convolutions in the shortcuts and first layer. The second configuration applies \gls{lgnn} to the main branch and the shortcuts in the resblocks but still ignores the first layer. The final configuration applies \gls{lgnn} to all layers.  

\begin{table}[htbp]
    \centering
    \begin{tabular}{|>{\centering\arraybackslash}p{0.8cm}|c|c|c|c|c|c|c|}
          \hline
          \multirow{2}{*}{\begin{tabular}{c}WRN\\size\end{tabular}} & \multirow{2}{*}{Regular} & \multicolumn{2}{c|}{Main Branch} & \multicolumn{2}{c|}{ResBlocks} & \multicolumn{2}{c|}{All}\\ \cline{3-8} & & Con & Dec  & Con & Dec & Con & Dec \\
          \hline
          \hline
          {\begin{tabular}{c}16-8\end{tabular}} & 78.87 & 78.90 & 78.94 & 79.08 & 79.10 & 79.07 &79.12\\
          \hline
          {\begin{tabular}{c}28-10\end{tabular}} & 80.62 & - & - & 80.88 & 81.08 & 80.99 & 80.81\\
          \hline
    \end{tabular}
    \caption{Accuracies (\%) for Wide ResNet}
    \label{tab:wrn}
\end{table}

The results on WRN are shown in table \ref{tab:wrn}. In the WRN-16-8 experiment, All \gls{lgnn} configurations performed better than the baseline, with the \textit{Main Branch} configuration performing slightly worse than the other two. The other \gls{lgnn} configurations have relatively similar performances with the highest performance having a $0.25\%$ increase compared to the baseline. The best median performance came from having a decreasing sigma where the difference between applying \gls{lgnn} to only the resblocks or all layers was fairly minimal. In the WRN-28-10 experiment, since applying \gls{lgnn} to only the main branch of the resblocks did not perform as well as the other two structures in WRN-16-8, those experiments were left out. The best median performance came from applying \gls{lgnn} to only the resblocks with a decreasing sigma which had a performance increase of $0.46\%$ increase.

Overall, while the performance increase was minimal, it scaled positively with network size. Our theory as to why \gls{lgnn} marginally improves performance is due to regularization. While we use weight decay already, our other experiments in section \ref{exp_first} suggests that the competitive learning from \gls{lgnn} may have a regularizing effect on the weights. Another reason why \gls{lgnn} could possibly have an effect on performance is that competitive learning could allow individual neurons to escape local minimums. In these situations, if a neuron is stuck in a local minimum, as long as one of its neighbours receives a large gradient, the neuron in question will receive a part of it and eventually escape the local minimum. The main focus of this work was to make \glspl{cnn} more understandable while having minimal impact on the learning which was achieved by not having any of the median accuracies drop using our method.

\subsection{Analysis of \gls{lgnn} on the First Layer of a Network} \label{exp_first}

\begin{figure}[htbp]
    \captionsetup{justification=raggedright}
     \centering
     \begin{tabular}{ccc}
         \multicolumn{2}{c}{
         \begin{subfigure}[b]{0.25\textwidth}
             \centering
             \includegraphics[width=\textwidth]{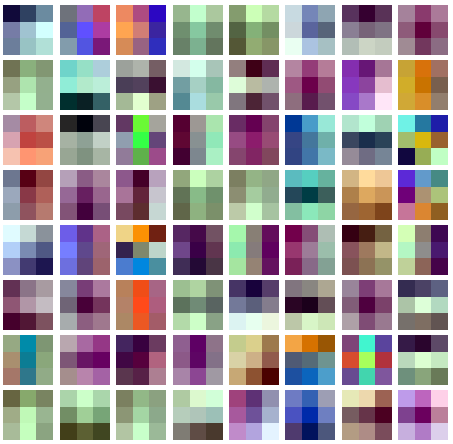}
             \caption{VGG-19 Baseline}
             \label{fig:reg_each}
         \end{subfigure}}
        \\
         \centering
         \begin{subfigure}[b]{0.25\textwidth}
             \centering
             \includegraphics[width=\textwidth]{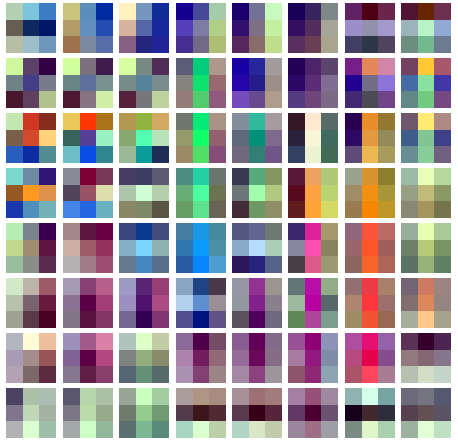}
             \caption{VGG-19 LGNN-Constant}
             \label{fig:lgnncon_each}
         \end{subfigure}
         &
         \begin{subfigure}[b]{0.25\textwidth}
             \centering
             \includegraphics[width=\textwidth]{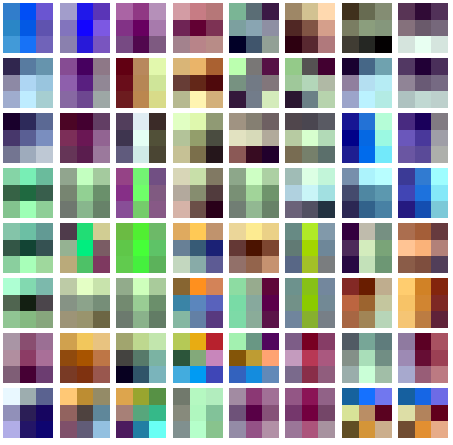}
             \caption{VGG-19 LGNN-Decreasing}
             \label{fig:lgnndec_each}
         \end{subfigure}
    \end{tabular}
    \caption{Comparing the first layer of VGG-19 between the baseline and \gls{lgnn}.}
    \label{fig:first_layers}
\end{figure}

\begin{table}[htpb]
    \centering
    \begin{tabular}{|c|c|c|c|}
        \hline
         Magnitudes & Regular & LGNN-con & LGNN-dec\\
         \hline\hline
         Min & $2.1775e-06$ & $0.0111$ & $0.0007$ \\
         \hline
         Max & $2.2977$ & $1.6394$ & $1.9456$\\
         \hline
         Std Dev of Log & $3.6854$ & $1.0199$ & $2.0902 $\\ 
         \hline
    \end{tabular}
    \caption{Magnitude statistics of filters in Figure \ref{fig:first_layers}}
    \label{tab:first_layers}
\end{table}

To further understand how \gls{lgnn} affects a network, we examine the filters for the first layer of VGG-19 for one set of trials. The first layer for VGG-19 is used because it is the only layer with 3 input channels which makes it easier to visualize and VGG has more filters in the first layer. For the other layers, visualization techniques in \cite{olah_feature_2017} are required. The filter weights were normalized between [0 1] individually and then organized to the \gls{som} dimensions. The first layer of VGG-19 contains 64 filters and the \gls{som} dimensions were [8,8].

As shown in figure \ref{fig:first_layers}, \gls{lgnn} has a noticeable effect on the learned filters. Compared to the baseline filters, the \gls{lgnn} filters have an organized structure where the filters with similar appearances are grouped together locally. Looking at table \ref{tab:first_layers}, the \gls{lgnn} filters also have a smaller range of magnitudes. In the table, the smallest magnitude for the baseline is in the range of $10^{-6}$ while the \gls{lgnn} with constant neighbourhood and decreasing neighbourhood have a minimum magnitude of $10^{-2}$ and $10^{-4}$. This smaller range of magnitudes points at our method having a regularizing effect that reduces overfitting. It is important to note that since batch normalization and weight decay were being used in training there already was some regularization, but \gls{lgnn} seems to still positively affect the training although minimally.

\subsection{Analysis of \gls{lgnn} on Hidden Layers of a Network}\label{exp_hidden}

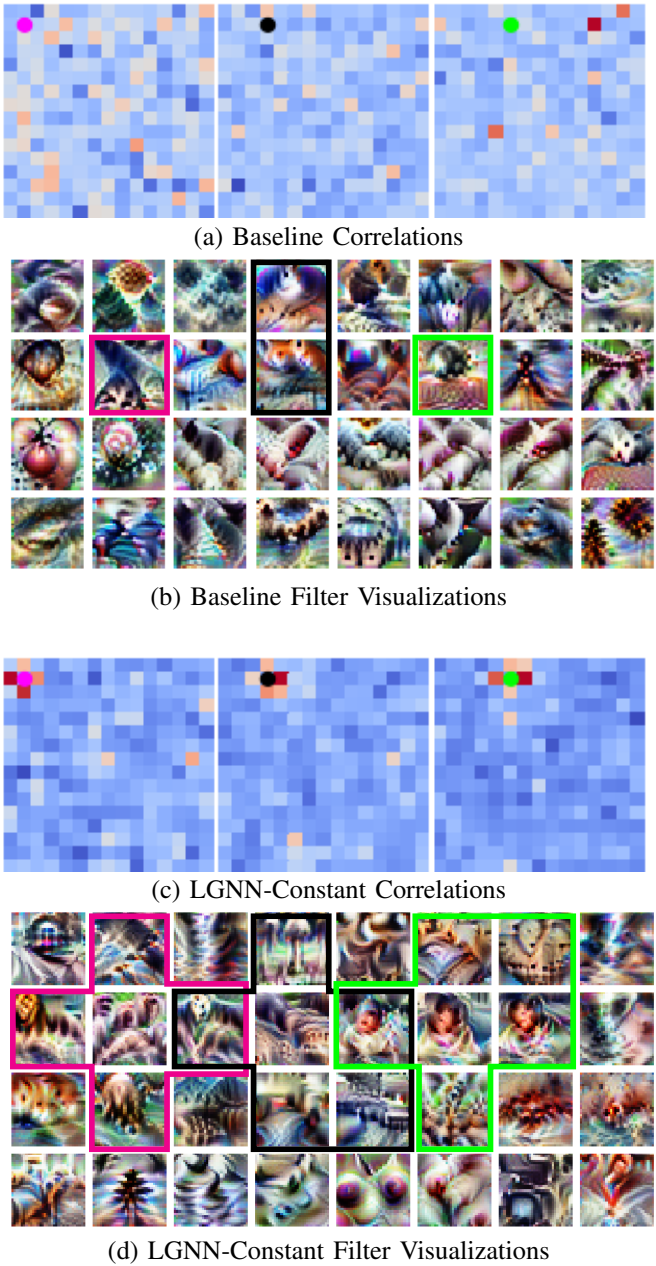
\begin{figure}[htbp]
    \begin{tikzpicture}
        \begin{scope}[xshift=0]
             \node{\includegraphics[trim={0.65cm 4.6cm 0cm 0.1cm},clip, width=0.5\textwidth]{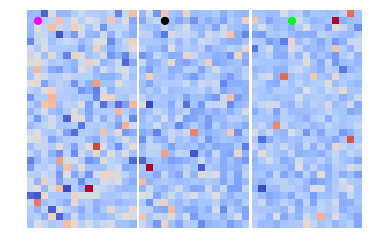}};
             \node[align=center, below] at (0,-1.5) {(a) Baseline Correlations};
        \end{scope}
        \begin{scope}[yshift=-4.1cm]
             \node{\includegraphics[trim={0cm 54cm 16cm 0cm},clip, width=0.48\textwidth, ]{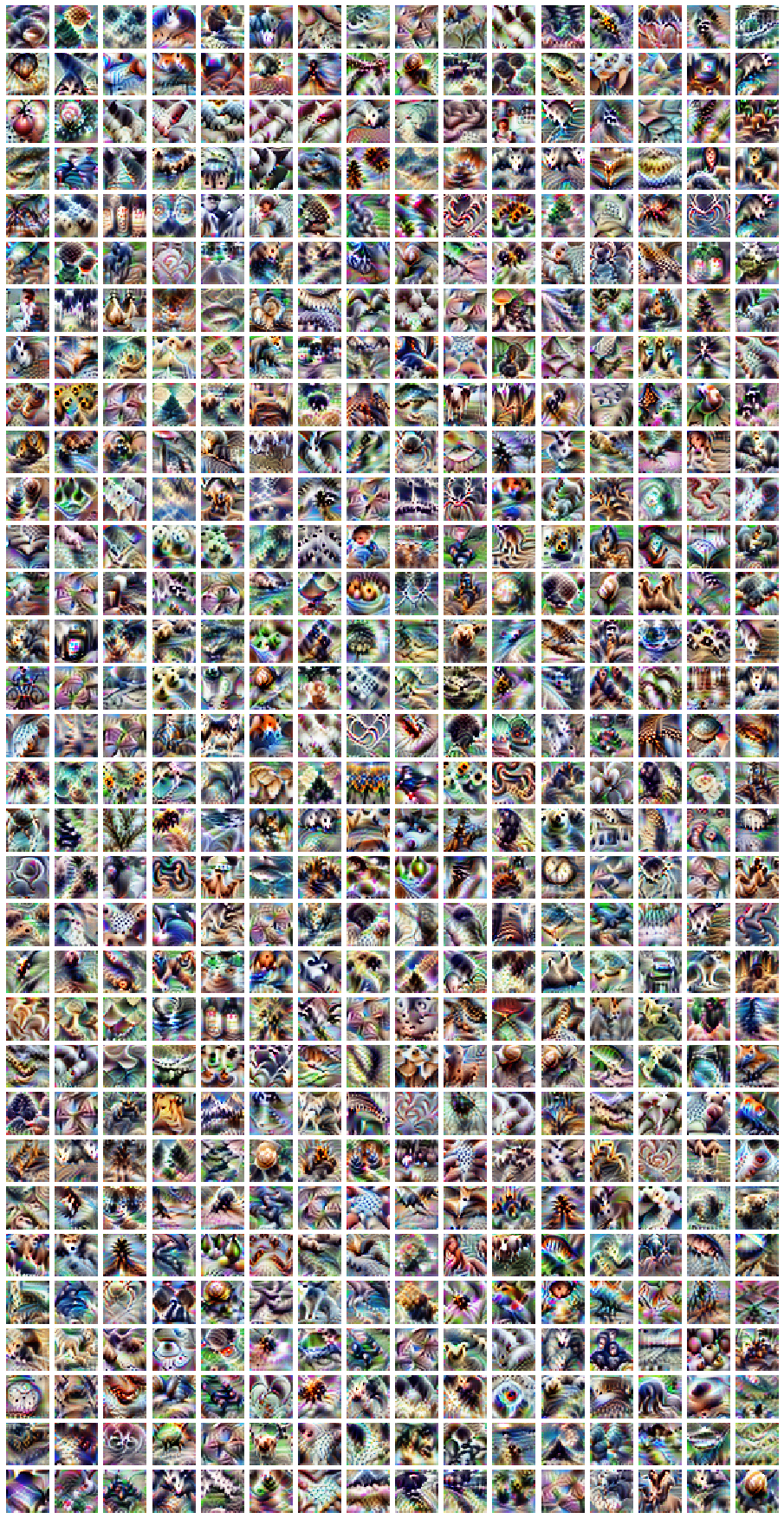}};
             \draw[line width=0.75mm, color=magenta, xshift=-2.15cm] (0,0) -- (0,1) -- (-1, 1) -- (-1, 0) -- cycle;
             \draw[line width=0.75mm] (0,0) -- (0,2) -- (-1, 2) -- (-1, 0) -- cycle;
             \draw[line width=0.75mm, color=green, xshift=2.15cm] (0,0) -- (0,1) -- (-1, 1) -- (-1, 0) -- cycle;
             \node[align=center, below] at (0,-2.2) {(b) Baseline Filter Visualizations};
        \end{scope}
        \begin{scope}[yshift=-8.7cm]
             \node{\includegraphics[trim={0.65cm 4.6cm 0cm 0.1cm},clip, width=0.5\textwidth]{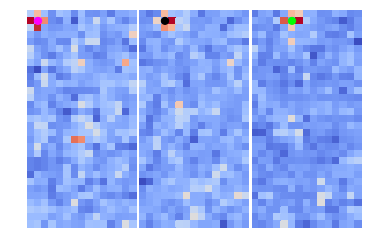}};
              \node[align=center, below] at (0,-1.5) {(c) LGNN-Constant Correlations};
        \end{scope}
        \begin{scope}[yshift=-12.8cm]
             \node{\includegraphics[trim={0cm 55.1cm 16cm 0cm},clip, width=0.48\textwidth, ]{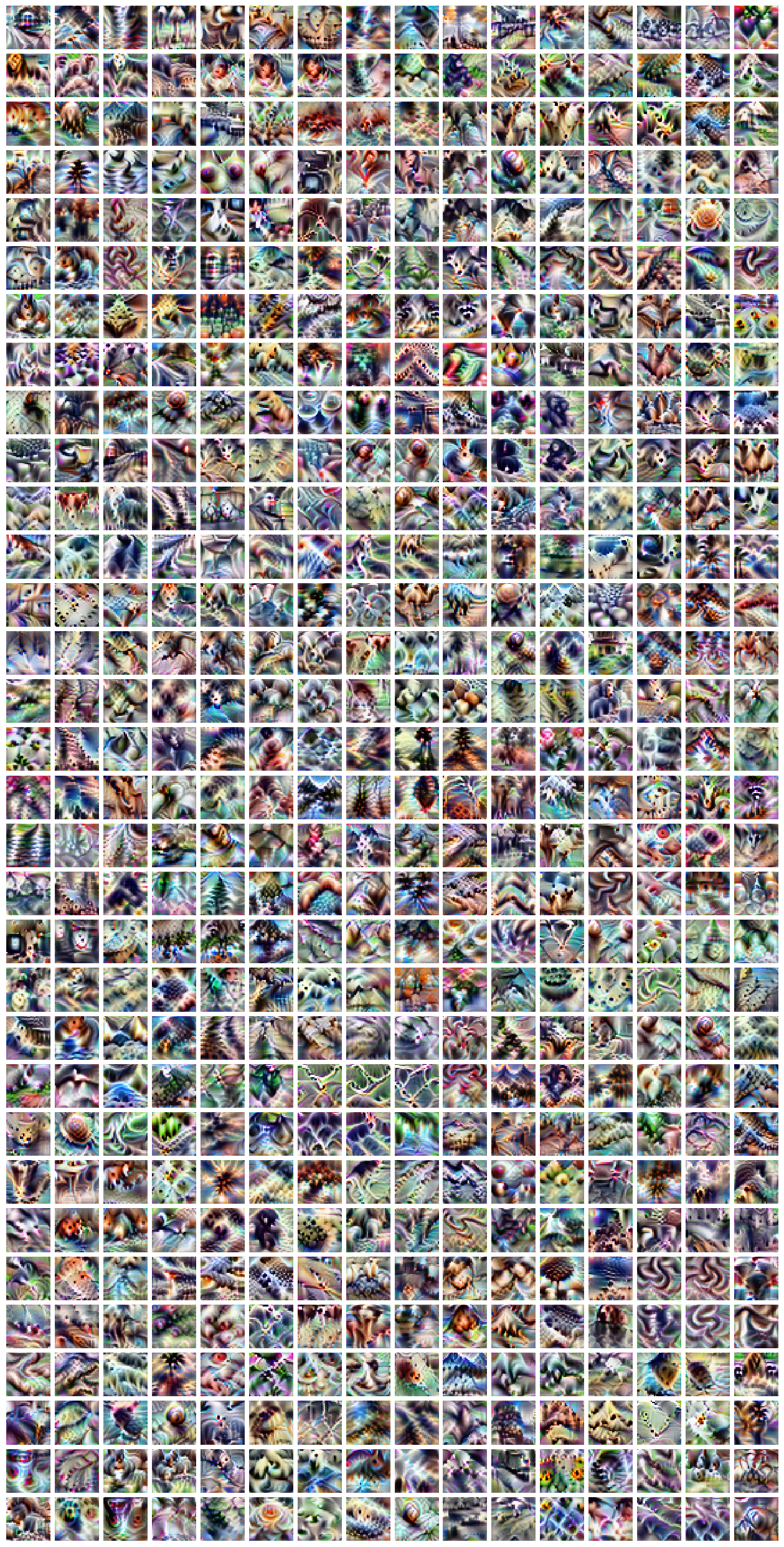}};
             \draw[line width=0.75mm, xshift=-2.15cm, color=magenta] (0,1.1) -- (0,2) -- (-1,2) -- (-1,1) -- (-2.05,1) -- (-2.05,0) -- (-1,0) -- (-1,-1.1) -- (0,-1.1) -- (0,-0.1) -- (1.05,-0.1) -- (1.05, 1.1) -- cycle;
             \draw[line width=0.75mm] (0,1) -- (0,2) -- (-1,2) -- (-1,1) -- (-2.05,1) -- (-2.05,0) -- (-1,0) -- (-1,-1.1) -- (1.1,-1.1) -- (1.1, 1) -- cycle;
             \draw[line width=0.75mm, xshift=2.15cm, color=green] (1.1,2) -- (-1,2) -- (-1,1.1) -- (-2.05,1.1) -- (-2.05,0) -- (-0.95,0) -- (-0.95,-1.1) -- (0,-1.1) -- (0,0) -- (1.1,0) -- (1.1, 1) -- cycle;
             \node[align=center, below] at (0,-2.2) {(d) LGNN-Constant Filter Visualizations};
        \end{scope}
    \end{tikzpicture}
    \caption{ Correlations between filters within the last layer of VGG-11.}
    \label{fig:correlations}
\end{figure}

\begin{figure*}[htbp]
    \begin{tikzpicture}[spy using outlines={rectangle,black, connect spies}]
        \begin{scope}
            \node{\includegraphics[width=0.2\textwidth]{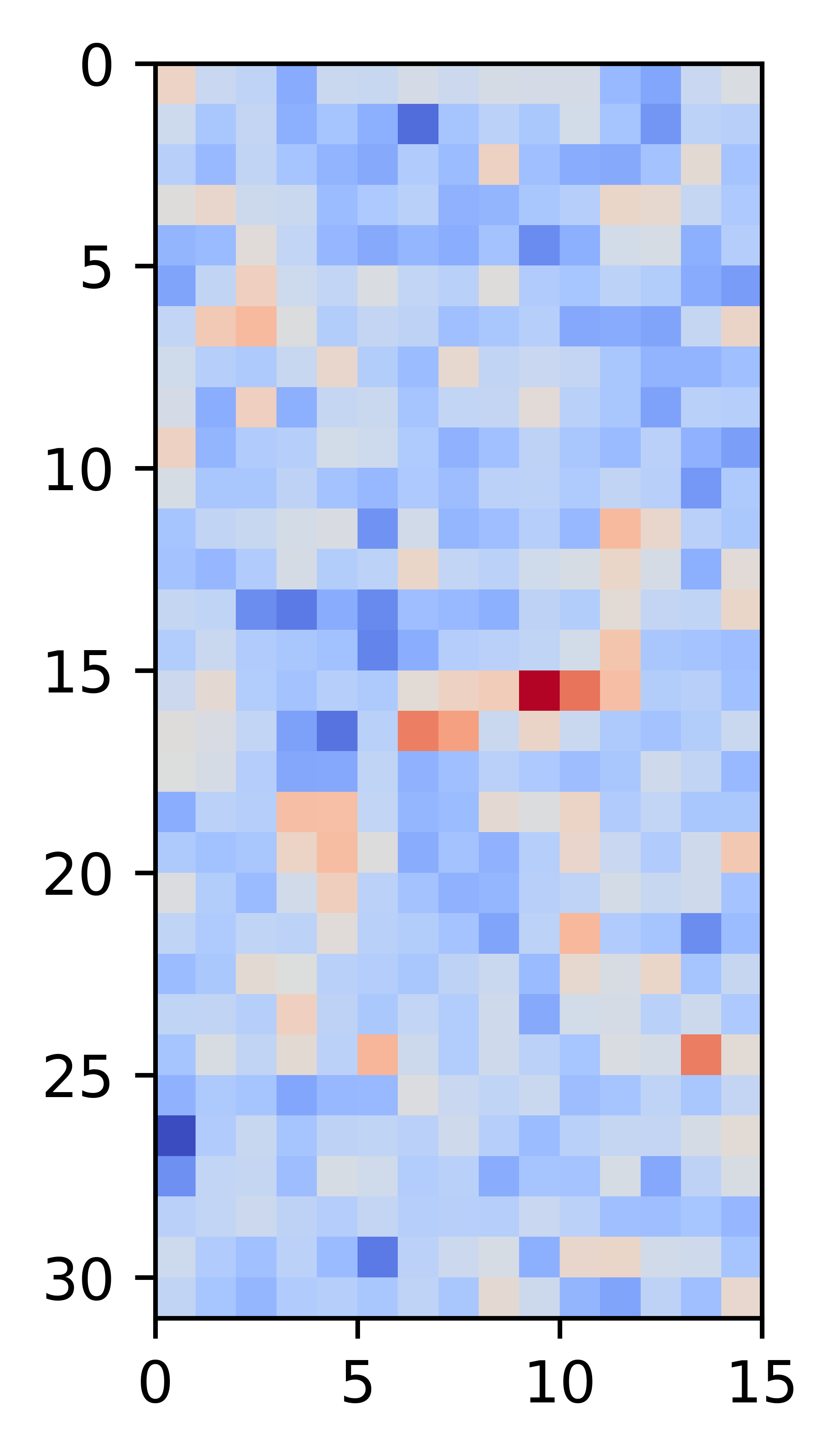}};
            \node[align=center, below] at (0,-3) {(a) Pine Tree};
            \spy[height=0.95cm,width= 2.45cm,magnification=2] on (0.45,0.1) in node [below] at (2.8,2); 
            \node[align=center, below] at (2.8,1) {(b)};
            \spy[height=0.95cm,width=0.95cm,magnification=2] on (-0.45,-0.45) in node [below] at (2.8,-1);
            \node[align=center, below] at (2.8,-2) {(c)};
        \end{scope}
        \begin{scope}[xshift=5.8cm]
            \node{\includegraphics[width=0.2\textwidth]{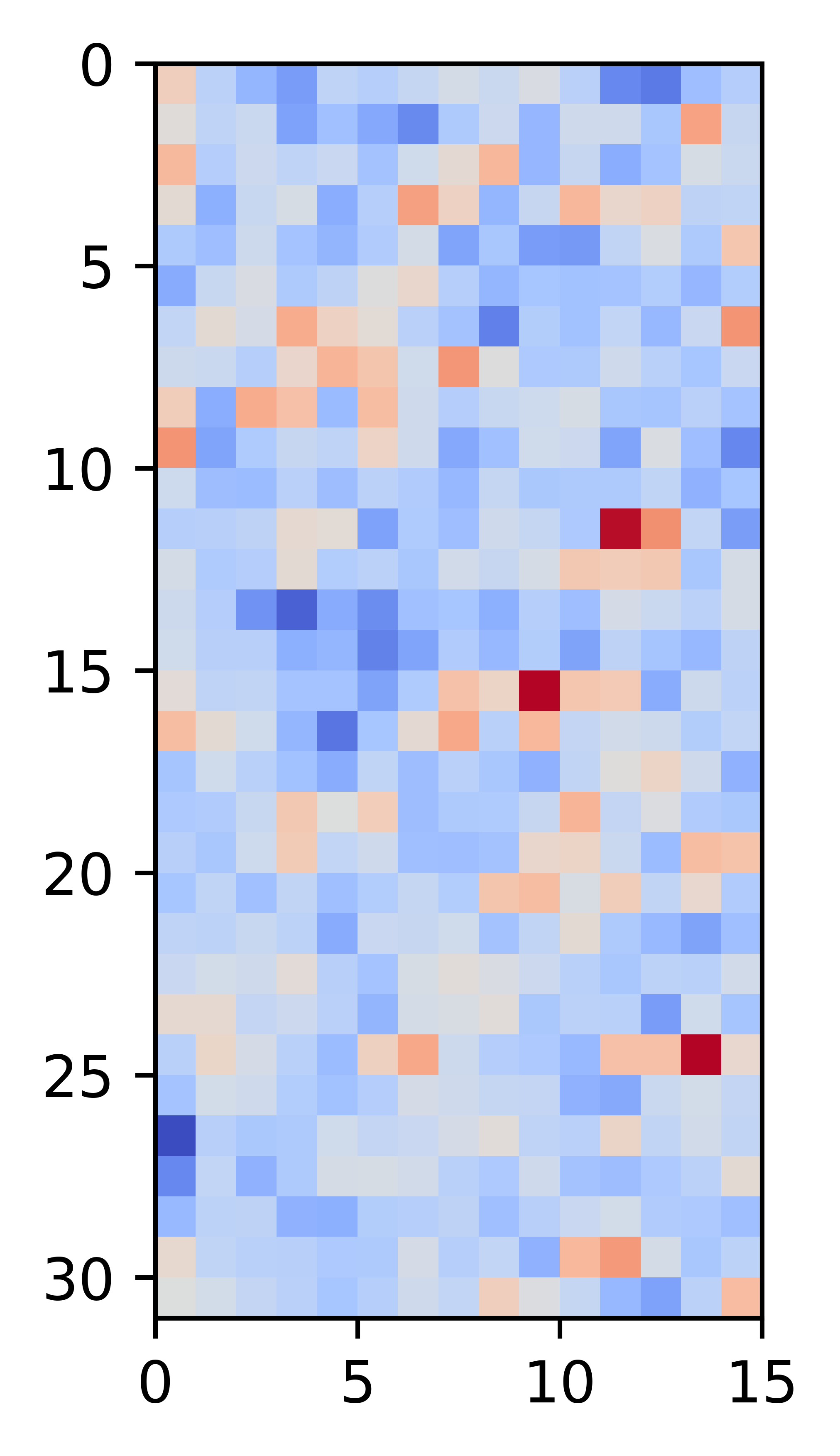}};
            \node[align=center, below] at (0,-3) {(d) Maple Tree};
            \spy[height=0.95cm,width= 2.45cm,magnification=2] on (6.25,0.1) in node [below] at (8.6,2); 
            \node[align=center, below] at (2.8,1) {(e)};
            \spy[height=0.95cm,width=0.95cm,magnification=2] on (5.35,-0.45) in node [below] at (8.6,-1);
            \node[align=center, below] at (2.8,-2) {(f)};
        \end{scope}
            \begin{scope}[xshift=11.5cm]
            \node{\includegraphics[width=0.2\textwidth]{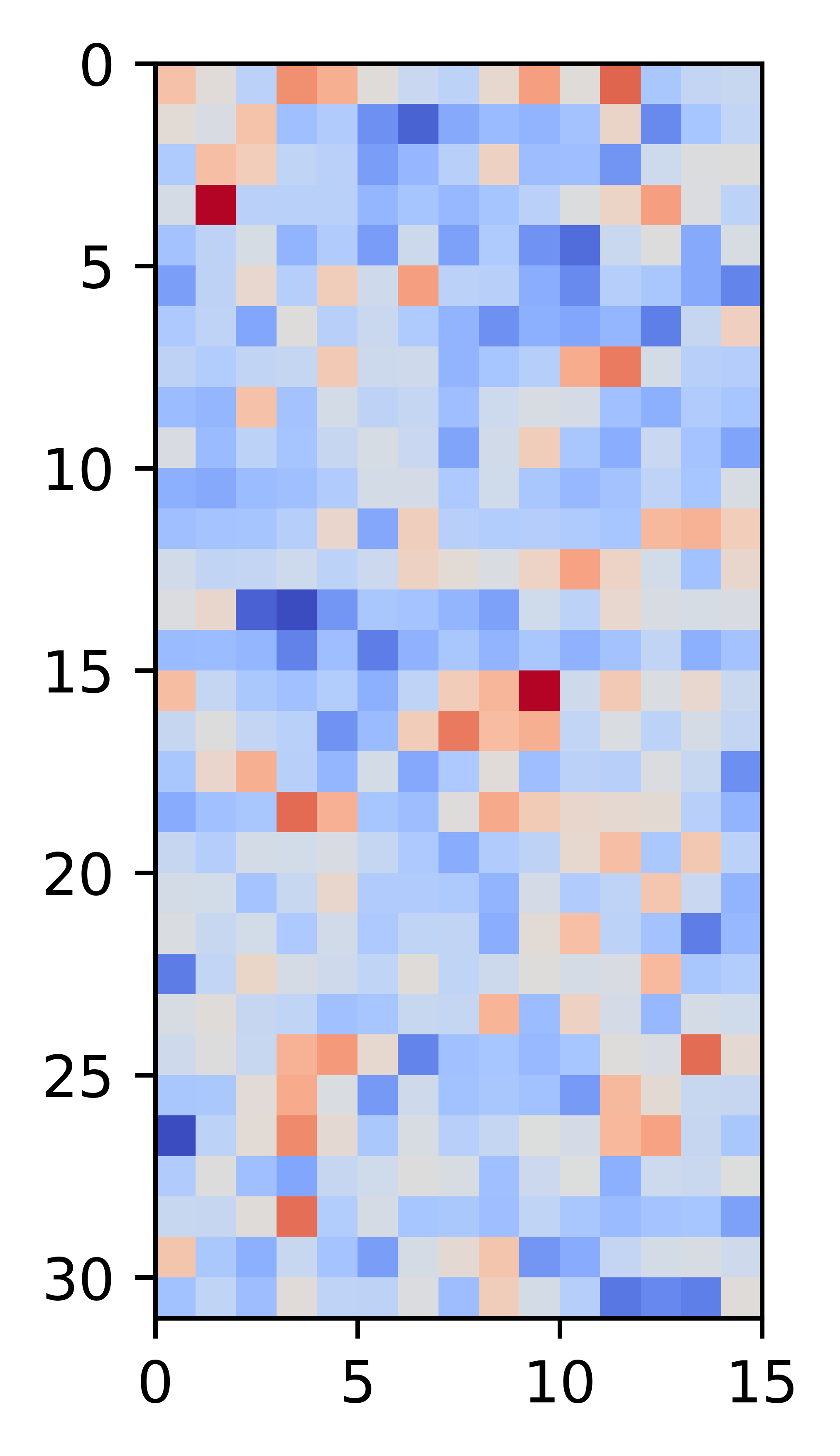}};
            \node[align=center, below] at (0,-3) {(g) Palm Tree};
            \spy[height=0.95cm,width= 2.45cm,magnification=2] on (11.95,0.1) in node [below] at (14.3,2); 
            \node[align=center, below] at (2.8,1) {(h)};
            \spy[height=0.95cm,width=0.95cm,magnification=2] on (11.05,-0.45) in node [below] at (14.3,-1);
            \node[align=center, below] at (2.8,-2) {(i)};
        \end{scope}
        \begin{scope}[yshift=-5.5cm, xshift=4cm]
            \node{\includegraphics[trim={14cm 29.5cm 6.1cm 29.5cm},clip, width=0.6\textwidth, ]{filters_Add_7_old.png}};
            \node[align=center, below] at (0,-2) {(j)};
        \end{scope}
        \begin{scope}[yshift=-5.5cm, xshift=12cm]
            \node{\includegraphics[trim={6.1cm 23.7cm 22.0cm 35.5cm},clip, width=0.2\textwidth, ]{filters_Add_7_old.png}};
            \node[align=center, below] at (0,-2) {(k)};
        \end{scope}
    \end{tikzpicture}
    \caption{Heat maps for average activations of the last convolution layer for 3 tree classes in CIFAR-100: Pine (a), Maple (d), and Palm (g) . For each class, activations were averaged over all 100 test images and then reshaped to \gls{som} dimensions (32,16). Filter clusters shown by (b,e,h) and (c,f,i) seem to activate together for tree images. (j) and (k) show visualizations from Google's Lucid toolbox for (b,e,h) and (c,f,i) respectively}
    \label{fig:activations}
\end{figure*}

While the first layer of a network is generally the only layer where weights can be directly mapped to color, the higher level layers are the ones that contain more interesting semantic filters. To visualize these filters, we use the techniques listed by \cite{olah_feature_2017} in which their code is available through Google's Lucid toolbox. The toolkit numerically generates images that maximally activates the filters. Our networks which were trained in Pytorch were transferred to Tensorflow using the Open Neural Network Exchange (ONNX) toolbox. Using the Lucid toolbox, the image size was set to $32 \times 32$, \textit{fft} and \textit{decorrelate} were set to \textit{True}, and all other parameters were kept at their default values.

The first analysis we perform on the higher levels is to observe if the topology holds and if these correlations correspond to clusters of filters with similar semantic information. To see if the correlations hold, we first calculate a Gram matrix for the filter weights of the last layer of VGG-11. Each row of the Gram matrix represents the correlation between the filter corresponding to that row and all the other filters. By extracting said row and reshaping it to the \gls{som} dimensions, we can then represent the correlations as a heat map. We set the main diagonal of the Gram matrix to 0 before extracting each row so the heat maps do not saturate. (a) and (c) of Figure \ref{fig:correlations} shows the heat maps of filters [1,1], [1,3], and [1,5]. We only show the top half  of the heat maps to save space.  Filter visualizations from Google's Lucid are also included in (b) and (d) to give readers a better understanding. Any correlated (red) neighbours in the heat maps have been outlined in the filter visualizations in their respective colors.

As shown in Figure \ref{fig:correlations}, the \gls{lgnn}-Constant version shows more correlations in a local area than the baseline version. In the heat maps for the baseline (a), most of the neighbours to each of the filters of interest are not highly correlated. Only the middle heat map has a single instance of a correlated neighbour. As for the heat maps for the \gls{lgnn} version (c), all the direct neighbours of the indicated filters are highly correlated as shown by the red '+' in the heat map. The heat map in the middle and right even has one of the diagonals slightly correlated. The reason its not a red square is due to the choice of our LPF weights that we used in \gls{lgnn}. Looking at (c) and (d), the filters in the outlined areas do indeed share some visual similarities, especially the filters outlined in green for (d) where the filters to the left and right are remarkably similar.

While the previous example demonstrates the correlation between neighbouring filters, the correlations may not necessarily be strong enough to assume that the filters will activate in clusters during inference. In Section \ref{back_visualization}, we discussed attribution methods as a means of seeing how filters react to individual images. Unfortunately, getting an idea on how clusters of filters would react on average to various examples would be difficult for attribution methods. Instead, we look at the average activations for a full layer for a batch of images in one class. Since we have already shown in Figure \ref{fig:correlations} that the baseline network does not hold any correlation between neighbouring filters, we only show the activations for \gls{lgnn}-Constant. We take the 100 test images for each class for 3 classes in CIFAR-100 and feed them into the network. We then take the activations for the last convolution layer and average over the spatial and batch dimensions. The average activations are then mapped to the \gls{som} dimensions and represented using a heat map.

In Figure \ref{fig:activations}, we first visualized all of the filters in the last layer, and found a cluster of filters that looked like trees shown in (k). Passing the classes (a) Pine Tree, (d) Maple Tree, and (g) Palm Tree we found that there were a few small positive activations in this cluster, but more importantly the region in (j) showed various filters in a larger cluster that also activated between various tree classes. Within the cluster in (j), one particular filter is the highest activation between the various trees while the other filters around it activate in differing proportions depending on the type of tree. It is important to note that the images fed into the network were the test images and not the training images. The test accuracies for these classes were 70\% for the pine tree, 67\% for the maple tree, and 90\% for the palm tree. While observing the differences between activation maps for the incorrect and correct outputs would be interesting, it is outside the scope of this paper. One important detail is that the clusters of similar looking images correspond to correlated activations on average, and also the fact that despite having different output labels, 3 different tree classes had various positive activations for a specific cluster within the \gls{lgnn} filters.        

\section{Conclusion}\label{conclusion}

Our proposed method for XAI, \gls{lgnn} is capable of gathering clusters of similar filters during training for a neural network. These clusters can often reveal to a user which filters respond to similar semantic concepts. Using the concept of competitive learning and neighbourhood functions which was inspired by \gls{som}, our back propagation allows neurons within the same layer to share information with each other. One great benefit is that our system only modifies the back propagation and leaves the model structure and forward inference in tact. This makes our method fairly attractive if users wish to use our method for transfer learning where the trained network is being used only as a feature detector and a separate classifier is being trained that could possibly be hand crafted to take advantage of the organized filter structure.  

Our experiments show reasonable success in making neural networks more explainable. In our first analysis we show that for different network types, the clustering from \gls{lgnn} does not impede the learning capacity of the network and in fact offers a small accuracy increase on average. We believe this small accuracy improvement is due to the neighbourhood functions being a good regularizer. The reason why the accuracy improvement is fairly minimal could be due to the fact that the baseline networks already contained batch normalization and weight decay as regularizers already. In our second analysis, we observe the effect on \gls{lgnn} within the first layer of a network. It was fairly evident that \gls{lgnn} was capable of enforcing a topology onto the filters where clusters of neighbouring filters shared similar properties. We also compare the range of the magnitudes of the filters and see that \gls{lgnn} does have a regularizing effect on the filters. In our last analysis, we examined the effect of \gls{lgnn} on the last layer. The correlations between neighbouring filters still held up for \gls{lgnn} in comparison to the baseline. We also managed to demonstrate that these correlations can lead to clusters that activate for similar semantic concepts, like how certain clusters activated for different types of trees even though the varieties of trees had different output labels.  

Several future improvements to the system can be made. First of all, although our system takes concepts from \gls{som}, we were not able to properly incorporate 'winner takes all' without it adversely affecting the learning. If winner takes all is reincorporated into the learning, it could possibly force similar filters to a singular cluster rather than several smaller clusters. Additionally, we can investigate tuning the hyper parameters for the neighbourhood function for each layer rather than sharing them between all layers.

\bibliographystyle{IEEEtran}

\bibliography{lgnn}
\end{document}